\documentclass[11pt]{article}

\usepackage[final]{acl}

\usepackage{ulem}

\usepackage{times}
\usepackage{makecell}
\usepackage{multirow}
\usepackage{latexsym}
\usepackage{natbib}
\usepackage{amsmath,amssymb,amsfonts}
\usepackage{graphicx}
\usepackage{subcaption}
\usepackage{textcomp}
\usepackage{xcolor}
\usepackage{url}
\usepackage{algorithm}
\usepackage{algpseudocode}
\usepackage{tabularx}
\usepackage{adjustbox}
\usepackage{booktabs}
\usepackage{tikz}
\usepackage{circuitikz}
\usepackage{enumitem}
\usepackage{wasysym}
\usepackage{colortbl}
\usepackage[most]{tcolorbox}
\usepackage[table]{xcolor}
\definecolor{cellbg}{RGB}{230, 245, 255} 
\definecolor{headerbg}{RGB}{230, 245, 255} 
\definecolor{red4}{HTML}{febf92}
\definecolor{dt}{gray}{0.7}
\def\BibTeX{{\rm B\kern-.05em{\sc i\kern-.025em b}\kern-.08em
    T\kern-.1667em\lower.7ex\hbox{E}\kern-.125emX}}
\usepackage{diagbox}        
\usetikzlibrary{shapes, arrows, positioning, calc} 
\tcbuselibrary{skins, breakable} 
\usepackage[T1]{fontenc}

\usepackage[utf8]{inputenc}

\usepackage{microtype}

\usepackage{inconsolata}

\usepackage{graphicx}

\title{VISTA: Mitigating Semantic Inertia in Video-LLMs via Training-Free Dynamic Chain-of-Thought Routing}

\author{
 \textbf{Hongbo Jin\thanks{Equal contribution}} \quad
 \textbf{Jiayu Ding\footnotemark[1]} \quad
 \textbf{Siyi Xie\footnotemark[1]} \quad
 \textbf{Guibo Luo} \quad
 \textbf{Ge Li\thanks{Corresponding author}}
\\
 School of Electronic and Computer Engineering,\\
 Peking University
\\
 \small{
   \textbf{Correspondence:} \href{hbjin25@stu.pku.edu.cn}{\{hbjin25, jyding25\}@stu.pku.edu.cn}
 }
}

\begin{document}
\maketitle
\begin{abstract}
Recent advancements in Large Language Models have successfully transitioned towards System 2 reasoning, yet applying these paradigms to video understanding remains challenging. While prevailing research attributes failures in Video-LLMs to perceptual limitations, our empirical analysis reveals a cognitive misalignment termed Semantic Inertia, where models suppress valid visual evidence in favor of dominant language priors. To rectify this, we propose VISTA, a training-free framework designed to align perception with logical deduction. By dynamically routing inference paths and materializing implicit visual features into explicit textual anchors, our approach effectively counterbalances the influence of parametric knowledge.
Furthermore, we incorporate a Latent Reasoning Consensus mechanism to mitigate stochastic hallucinations. VISTA  showed outstanding results on a wide range of benchmarks, and outperforms its base model by 9.3\% on Egochema and 5.6\% on VideoEspresso, rivalling or even surpassing larger and proprietary models. Our codebase will be publicly available soon.
\end{abstract}

\section{Introduction}
The evolution of Large Language Models (LLMs) towards System 2 reasoning, driven by Chain-of-Thought (CoT), has successfully elevated the text processing paradigm from shallow pattern matching to explicit logical deduction, significantly enhancing robustness in complex tasks~\cite{li2025system,jaech2024openai,deepseekai2025deepseekr1incentivizingreasoningcapability}. Aligning with this trend, recent works have introduced System 2 paradigms into video understanding, attempting to improve end-to-end Video-LLMs via CoT.
Mainstream approaches~\cite{zhang2025video, wen2025agricot, muennighoff2025s, ye2025limoreasoning} typically employ Supervised Fine-Tuning (SFT) with video CoT instruction data, and recently incorporate Reinforcement Learning (RL) techniques~\cite{wang2025videorft, deepseekai2025deepseekr1incentivizingreasoningcapability} to further enhance reasoning reliability and alignment.
However, this data-driven alignment fails to alter the model's black-box nature: visual information remains encapsulated as implicit latent embeddings, lacking observable intermediate representations. This raises a critical question: when models fail in complex video reasoning, does the bottleneck lie in \textit{perception} or \textit{reasoning}?

Prevalent academic views typically attribute these failures to perceptual limitations, assuming that visual encoders fail to effectively extract complex spatiotemporal features. Consequently, substantial research~\cite{wang2024internvideo2,ren2024timechat} has been dedicated to scaling up visual encoders or introducing fine-grained spatiotemporal modules, aiming to improve performance by enhancing perceptual fidelity. However, our probe experiments offer contrary empirical evidence.
Even in instances where the model fails to answer complex queries, it maintains high accuracy on the underlying atomic visual questions that support these complex reasoning tasks.
This finding confirms that key atomic visual facts remain encoded within the latent representations.
This implies that the absence or inaccuracy of visual representations may not be the dominant cause of failure; instead, the issue likely stems from their ineffective utilization during the subsequent language generation process.

We identify this as a deep cognitive misalignment, which we conceptually frame as ``\textbf{Semantic Inertia}''. From the perspective of autoregressive generation, the model must balance intrinsic language priors with extrinsic visual context. Our observations suggest that the strong parametric knowledge acquired during massive text pre-training often overwhelms visual input. Consequently, the model tends to ignore visual constraints, prioritizing generation paths driven by statistical text patterns. This implies that hallucinations stem less from perceptual failures than from the dominance of language priors over visual evidence.

To mitigate this ``Semantic Inertia'' and address the perception-reasoning misalignment, we propose \textbf{V}isual \textbf{I}nference via \textbf{S}tructured \textbf{T}ext \textbf{A}nchoring (\textbf{VISTA}). This training-free framework shifts the paradigm from prior-driven generation to evidence-grounded deduction through three synergistic modules: (1) \textit{Dynamic Inference Routing} bypasses the pitfalls of  semantic inertia by intercepting complex queries and diverting them away from shallow statistical shortcuts; (2) \textit{Explicit Visual Anchoring} transforms implicit latent features into explicit textual descriptions, effectively materializing visual evidence to counterbalance the dominance of language priors; and (3) \textit{Latent Reasoning Consensus} serves as a logical verifier, filtering out stochastic hallucinations induced by language priors through a multi-path consensus mechanism.

This systematic paradigm effectively unlocks the model's potential perceptual capabilities. Validated on benchmarks including EgoSchema, VideoEspresso, VideoMMMU, MVBench, and PerceptionTest, VISTA achieves superior performance without any parameter updates. Notably, it improves accuracy by 9.3\% on EgoSchema and 5.6\% on VideoEspresso, rivaling or surpassing larger closed-source models like GPT-4o and Gemini-1.5-Pro.
Our contributions are summarized as follows:
\begin{itemize}
	\item We identify ``Semantic Inertia'' as a critical bottleneck in Video-LLMs, revealing that reasoning failures primarily stem from the suppression of valid visual evidence by dominant language priors rather than intrinsic perceptual limitations.
	\item We propose VISTA, a novel training-free System 2 reasoning framework designed to mitigate this perception-reasoning misalignment. It achieves this by dynamically routing inference paths and explicitly anchoring reasoning to materialized visual evidence.
	\item Extensive experiments on multiple video understanding benchmarks demonstrate that VISTA achieves competitive performance, comparable to or exceeding larger closed-source models, validating the effectiveness of our approach in video reasoning.
\end{itemize}

\section{Pilot Experiments}

To validate the ``Semantic Inertia'' hypothesis, we conducted a controlled pilot probe experiment. This study aims to address three pivotal questions: First, do failures in complex video reasoning stem primarily from perceptual deficiencies? Second, does the model possess the necessary reasoning capabilities when visual information is explicitly provided? Third, if both perceptual and reasoning modules are functional, what mechanism causes the model to suppress visual evidence in favor of language priors?

\subsection{The Perception-Reasoning Gap}

To decouple perception from reasoning, we curated 100 hard negative samples from MVBench where LLaVA-Video failed. Utilizing an annotation pipeline (detailed in Appendix), we extracted the Atomic Visual Facts (AVFs) essential for answering these queries.

\begin{figure}[t]
	\centering
	\includegraphics[width=0.95\linewidth]{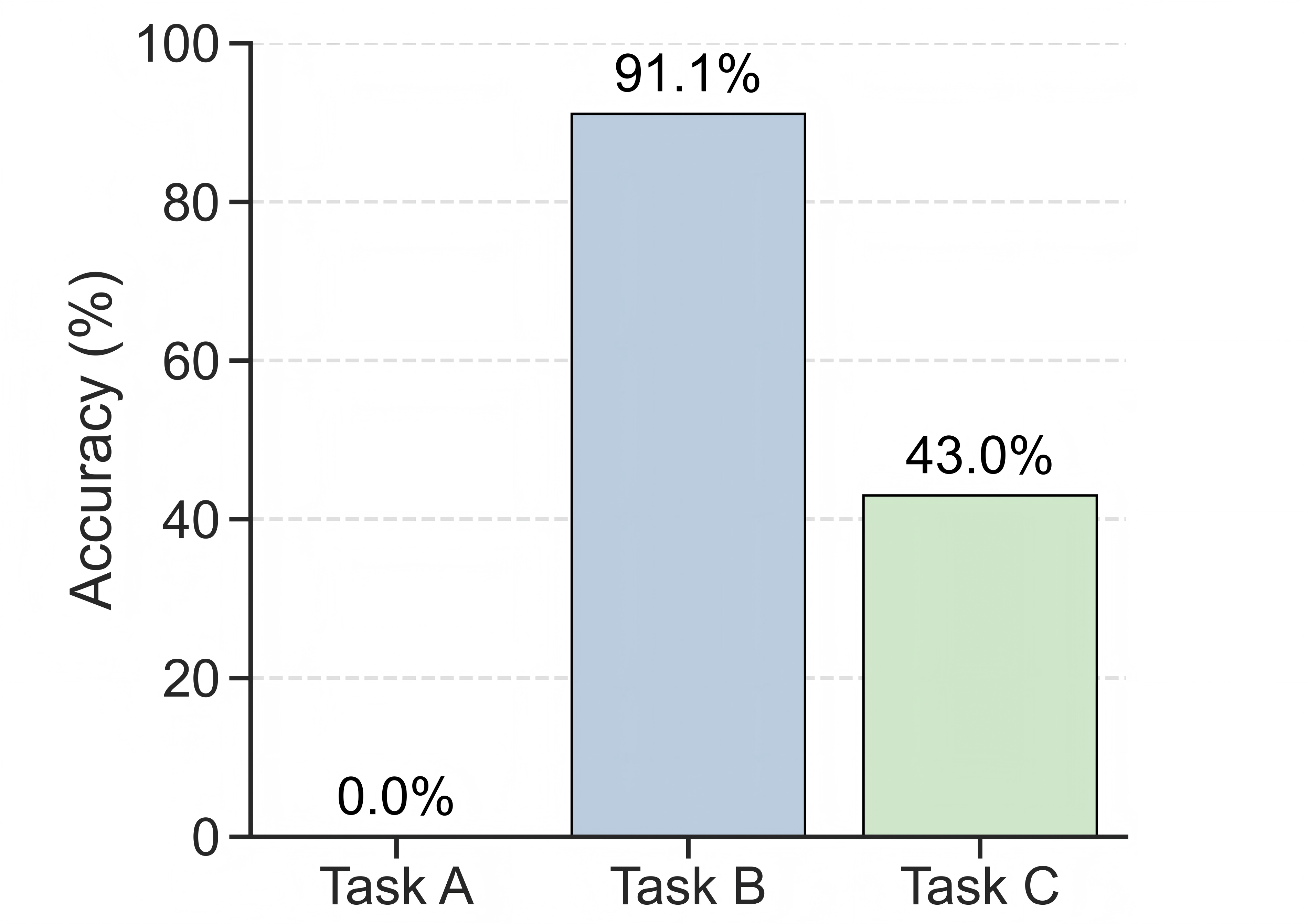}
	\caption{Performance Discrepancies across tasks.}
	\label{fig:pilot}
\end{figure}

\textbf{Settings and Results.} We established three progressive tasks: 1) \textbf{Task A:} The model performs standard end-to-end reasoning using the original video and complex queries. 2) \textbf{Task B:} Specific probe questions targeting AVFs are posed to validate whether visual features are accurately perceived by the model. 3) \textbf{Task C:} We convert the visual facts correctly identified in Task B into textual context and input them directly into the model (bypassing visual processing) to re-evaluate the complex queries from Task A.
The results in Figure \ref{fig:pilot} reveal significant performance disparities. Within the same sample set where Task A yields near 0\% accuracy (by design), Task B achieves an impressive \textbf{91.1\%}, indicating that the visual encoder successfully captures key spatiotemporal details. Crucially, Task C accuracy surges to \textbf{43.0\%}. This substantial recovery demonstrates that the model's logical reasoning module is capable when provided with correct premises.

\textbf{Mechanism Analysis.} The contrast between Task A and Task B provides strong evidence that the failure does not stem from the visual encoder's inability to capture essential spatiotemporal features. Simultaneously, the performance gap between Task A and Task C effectively rules out the possibility that the language module inherently lacks the requisite logical reasoning capabilities. This triangular verification pinpoints the primary failure mode as a ``Perception-Reasoning Misalignment'': despite being successfully encoded, the visual cues suffer from functional silence, as the logical module fails to effectively access or leverage them during standard reasoning tasks.

\begin{tcolorbox}[
	colframe=black,              
	colback=gray!10,             
	sharp corners=southwest,     
	enhanced,                    
	breakable,                   
	boxrule=1pt,                 
	width=\linewidth             
	]
	\textbf{Main Finding 1:} The primary bottleneck in current Video-LLMs is the ineffective utilization of cross-modal information. Despite being successfully encoded, visual facts remain in a state of ``Functional Silence'' and are not actively involved in the reasoning process as valid premises.
\end{tcolorbox}

\subsection{Semantic Inertia Suppression}

To determine why visual evidence is ignored, we investigated whether errors in Task A stem from random noise or systematic bias by designing a blind consistency test.

\textbf{Settings and Results.} We re-evaluated the 100 failed samples from Task A using black frames to block visual input, eliciting ``blind guesses'' driven solely by language priors. Leveraging the multiple-choice format, we assessed option consistency by comparing the predicted choice index of the blind guess with the original error. 
Results show that the model selected the identical option in 61\% of cases, significantly surpassing the 25\% random baseline inherent to 4-option tasks ($\chi^2$ test, $p < 0.001$, , indicating non-random systemic alignment). This suggests that the output distribution is dominated by language priors and remains largely invariant to video input.

\textbf{Mechanism Analysis.} This provides statistical evidence supporting the Semantic Inertia Hypothesis. Formally, let $Q$ denote the question and $A$ the predicted answer. During end-to-end reasoning, the model's strong parametric priors ($\mathcal{K}_{\text{param}}$) tend to overwhelm the immediate visual context ($V$). Consequently, the inference degenerates to $P(A | Q, V) \approx P(A | Q, \mathcal{K}_{\text{param}})$. Although visual evidence $V$ exists (confirmed by Task B) and the logical path is viable (confirmed by Task C), the reasoning bypasses visual constraints, collapsing into generation paths dominated by text statistical priors.

\begin{tcolorbox}[
	colframe=black,
	colback=gray!10,
	sharp corners=southwest,
	enhanced,
	breakable,
	boxrule=1pt,
	width=\linewidth
	]
	\textbf{Main Finding 2:} In failure cases, the dominance of language priors causes the model to behave as a text-only generator by suppressing visual evidence. This implies that addressing the issue requires interventions that disrupt these priors, forcing the model to abandon blind guessing in favor of evidence-grounded reasoning.
\end{tcolorbox}

\begin{figure*}[t]
	\centering
	\includegraphics[width=\textwidth]{ov.jpg}
	\caption{
    Overview of the VISTA framework. The pipeline begins with Dynamic Inference Routing, where the Taxonomy Routing module references the external Taxonomy feature table to identify the query type (e.g., causal reasoning). For complex queries, the Visual Materializer translates visual cues into a Structured Visual Summary. This output supports the Latent Reasoning Consensus stage, which filters hallucinations through multi-path sampling and similarity-based clustering. Finally, the selected reasoning path is fused into Aggregated Comprehensive Evidence, guiding the Video Large Language Model to generate the final response.}
	\label{fig:overall pipeline}
\end{figure*}

\section{Method}
To mitigate the ``Semantic Inertia'' problem where internal language priors suppress external visual evidence, we propose \textbf{VISTA}. As a training-free framework, VISTA explicitly aligns perception with reasoning through a step-by-step paradigm. As illustrated in Figure \ref{fig:overall pipeline}, the framework comprises three synergistic stages:
(1) \textit{Dynamic Inference Routing} (Sec. \ref{sec:routing}), which circumvents the risks of heuristic processing by diverting complex queries away from shallow statistical shortcuts; (2) \textit{Explicit Visual Anchoring} (Sec. \ref{sec:anchoring}), which counterbalances dominant language priors by materializing implicit latent features into explicit textual evidence; and (3) \textit{Latent Reasoning Consensus} (Sec. \ref{sec:consistency}), which filters out stochastic hallucinations induced by these priors via a multi-path consensus mechanism.

\subsection{Dynamic Inference Routing}
\label{sec:routing}

Video understanding tasks exhibit distinct susceptibilities to Semantic Inertia. While atomic perception relies on explicit visual cues, complex reasoning is highly vulnerable to the dominance of language priors, where models tend to bypass visual evidence in favor of shallow statistical shortcuts. Consequently, uniform processing risks handling reasoning-intensive queries via this default heuristic mode, exacerbating prior-driven hallucinations. To mitigate this, we design a routing mechanism that functions as a cognitive gatekeeper, intercepting high-risk queries and diverting them from heuristic paths to ensure rigorous visual grounding.

\textbf{Taxonomy Construction.} 
We first construct a representative taxonomy by drawing a subset from challenging benchmarks, including VideoEspresso~\cite{han2025videoespresso}, LongVideoBench~\cite{wu2024longvideobench}, and EgoSchema. We employ GPT-4o~\cite{hurst2024gpt} as a knowledge distiller to analyze these samples and extract common linguistic patterns and syntactic structures. The distilled features are organized into a taxonomy chart and refined through manual verification. This process yields typical categories such as \textit{fact-retrieval}, \textit{process-description}, \textit{causal-reasoning}, and \textit{behavior-inference}. Detailed definitions are provided in the appendix.
\begin{figure*}[h]
	\centering
	\includegraphics[width=0.9\textwidth]{ov2.jpg}
	\caption{Verification methods overview. We show the implementation and core differences between four different verification mechanisms.}
	\label{fig:verification_methods}
\end{figure*}

\textbf{Taxonomy-Guided Routing.} 
Based on this taxonomy, we designate specific categories (e.g., causal reasoning, interaction analysis) as requiring the VISTA deep reasoning branch. During inference, we match the input question against the lexical features in our table. If a match is found for a complex category, the model activates the deep reasoning pipeline; otherwise, it follows a direct inference path. In cases of conflict where a question matches multiple categories, we adopt a policy of selecting the category with the highest number of keyword hits. This policy minimizes categorization ambiguity, ensuring that complex reasoning tasks are accurately identified and not mistakenly routed to the shallow branch.

\subsection{Explicit Visual Anchoring}
\label{sec:anchoring}
To counteract the ``Semantic Inertia'' where internal language priors suppress external visual evidence, we propose Explicit Visual Anchoring. This module fundamentally alters the inference structure by decomposing the generation process, forcing the model to explicitly acknowledge visual facts before drawing conclusions. This is achieved through two synergistic phases: \textit{Visual Evidence Materialization} and \textit{Evidence-Grounded Deduction}.

\textbf{Visual Evidence Materialization.}
To reverse the dominance of language priors, we must elevate visual signals from implicit latent embeddings to explicit tokens. In this phase, we guide the model to generate a structured description of the video content relevant to the query. This process materializes dormant visual facts into textual evidence, ensuring that perceptual information possesses sufficient context density to compete with parametric priors in the subsequent generation.

\textbf{Evidence-Grounded Deduction.}
To further integrate the global context and utilize the materialized facts, we guide the model to reinterpret the query through a deliberate reasoning process. Specifically, we append the standard CoT trigger ``Let us think step by step'' to the prompt. This acts as a cognitive activator, stimulating the generation of an initial reasoning chain and increasing the distribution of logic-driven long contexts. Finally, we synthesize the original video frames, the user query, the materialized summary, and the generated reasoning chain as the input. By conditioning the final generation on this comprehensive evidence, we ensure the answer is a logical conclusion anchored in visual facts rather than a product of language priors.

\subsection{Latent Reasoning Consensus}
\label{sec:consistency}
Following Explicit Visual Anchoring, it is essential to ensure that the model rigorously adheres to the materialized video evidence instead of reverting to language priors. This step is critical for minimizing hallucinations and enhancing generalization. To this end, we designed four verification methods: Naive Prompting, Majority Voting, Best-of-N, and Latent Reasoning Consensus, as illustrated in Figure \ref{fig:verification_methods}.

\textbf{Naive Prompting}, as the name suggests, relies solely on prompting templates to complete the verification of the previous text.
There are limitations that explicit prompts are perceived by the model as having a certain affective tendency, i.e., it tends to find a way to negate his previous answer regardless of whether the previous output was correct.

\textbf{Majority Voting} is to have the model sample the output multiple times and select the plurality of the answers as the final response.
This approach simply considers the final answer and ignores the intermediate process of reasoning, leading to a limited validation effect.

\textbf{Best-of-N Sampling} denotes setting up a standard for confidence calculation and ultimately choosing the answer with the highest confidence score from the sampled outputs.
We propose to calculate the confidence scores of the candidate outputs based on semantic similarity and select the highest score as the final output:
\begin{equation}
	\label{eq:bon similarity}
	\text{sim}_{\text{semantic}} = \frac{1}{2} \left( \cos(\mathbf{v}_q, \mathbf{v}_s) + 1 \right)
\end{equation}
Here, $\mathbf{v}_q$ and $\mathbf{v}_s$ represent the vectors obtained from the question and summary text segments.
\begin{algorithm*}[t]
	\caption{Similarity-based Clustering Algorithm}
	\label{alg: clustering}
	\begin{algorithmic}[1]
		\State $ C \gets \emptyset$ \Comment{$C$ means all cluster set, initialized as $\emptyset$}
		\For{ $t_i \in T$ } \Comment{$t_i$ and $T$ respectively means current pending text and all texts}
		\State $\text{added\_to\_cluster} \gets \text{False}$
		\For{$\text{cluster} \in C$}
		\If{$ \sigma(t_i, r_c) \geq \theta$}
		\Comment{$\sigma$ refers to the similarity formula \ref{eq:LRC sim}, $\theta$ is similarity threshold}
		\State $\text{cluster}.\text{append}(t_i$) \Comment{ a hyperparameter we manually set ($\theta \in [0,1]$)}
		\State $\text{added\_to\_cluster} \gets \text{True}$
		\State \textbf{break} \Comment{Next Clustering Loop}
		\EndIf
		\EndFor \Comment{$r_c$ means the representative element of current cluster}
		\If{$\neg \text{added\_to\_cluster}$} \Comment{We choose $r_c = c[0]$ (i.e., the first text added)}
		\State $C.\text{append}([t_i])$
		\EndIf
		\EndFor
	\end{algorithmic}
\end{algorithm*}

\textbf{Latent Reasoning Consensus},
designed to enforce logical rigor, ensures that diverse reasoning paths converge to a unified semantic conclusion. Unlike character-level metrics (e.g., ROUGE or LCS) that are sensitive to lexical variations, we coalesce sampled paths based on their deep semantic alignment.
We calculate the reasoning consistency score by:
\begin{equation}
  S(r_i, r_j) = \frac{\mathbf{e}_i \cdot \mathbf{e}_j}{\|\mathbf{e}_i\| \|\mathbf{e}_j\|} 
    \label{eq:LRC sim}
\end{equation}
Here, $\mathbf{e}_i$ and $\mathbf{e}_j$ represent the last hidden state embeddings of the
final token of sampled reasoning paths $r_i$ and $r_j$, respectively.
This metric captures intrinsic logical agreement in the high-dimensional latent space rather than surface-level text overlaps.
Paths exceeding the semantic similarity threshold are clustered following Algorithm \ref{alg: clustering}.



\begin{table*}[t]
	\centering
	\small
	\setlength{\tabcolsep}{4pt} 
	\newcolumntype{Y}{>{\centering\arraybackslash}X}
	
	\begin{tabularx}{\textwidth}{l c YYYYY}
		\toprule[1.pt]
		\textbf{Model} &\textbf{Frames}& \textbf{ES} & \textbf{MV} & \textbf{PT} & \textbf{VE} & \textbf{VM} \\
		\midrule
		\multicolumn{7}{l}{\emph{\textbf{Proprietary Models}}} \\ 
		GPT-4V \cite{achiam2023gpt} &64& - &   43.5 & - & -&- \\
		GPT-4o \cite{openai_gpt4o_2024} &64& -    & - & - & -&61.2 \\
		Gemini-1.5-Flash \cite{google_gemini_1.5_2024} &128& 65.7 &   - & - & 39.8&49.8 \\
		Gemini-1.5-Pro \cite{team2023gemini} &128& 72.2 &   - & - & 44.2&53.9\\
		\hline
		\multicolumn{7}{l}{\emph{\textbf{Open-Source Models}}} \\
		LLaMA-VID-7B \cite{li2024llamavid}& 1fps &38.5&41.4&-&-&- \\
		LLaVA-Mini-8B \cite{zhang2025llavamini}& 1fps &51.2&44.5&-&-&- \\
		LLaVA-interleave-7B \cite{li2025llavanextinterleave}&- &-& 53.1&-&-&-\\
		TS-LLaVA-34B \cite{qu2024ts} &16& 57.8 &   - & - & -&- \\
		VILA-40B \cite{liu2025nvila} &256& 58.0 &   - & 54.0 & -&34.0 \\
		PLLaVA-34B \cite{xu2024pllava} &16& - &   58.1 & - & -&- \\
		LongVA-7B \cite{zhang2024long} &128& - &   - & - & 39.7&24.0 \\
		VideoLLaMA2-7B \cite{damonlpsg2024videollama2}  &16&53.3&53.9&52.2&-&-\\
		VideoLLaMA2-72B &16&63.9&62.0&57.5&-&- \\
		IXC-2.5-7B \cite{zhang2024internlm} &-&    & 69.1 & 34.4 & -&- \\
		VideoChat2-8B \cite{li2024mvbench} &16&55.8&60.3&53.0&-&-\\
		LLaVA-OV-72B \cite{li2024llava} &32& 63.9 &   59.4 & 66.9 & 63.2&-\\ 
		LLaVA-Video-72B \cite{zhang2024video} &32& - &   - & - & 66.3&49.7 \\
		\hline
		\multicolumn{7}{l}{\emph{\textbf{Our Models}}} \\
		LLaVA-OV-7B &32& 60.1 &   56.1* & 57.1 & 44.0&33.9 \\
		\rowcolor{headerbg} 
		\textbf{LLaVA-OV-7B+VISTA} &32& \textbf{67.8}~\textcolor{red}{(7.7\textuparrow~)} &  \textbf{58.6}~\textcolor{red}{(2.5\textuparrow~)} & \textbf{62.4}~\textcolor{red}{(5.3\textuparrow~)} & \textbf{47.9}~\textcolor{red}{(3.9\textuparrow~)} & \textbf{38.6}~\textcolor{red}{(4.7\textuparrow~)} \\
		LLaVA-Video-7B &32& 57.3   & 58.6 & 67.9 & 48.8&34.4* \\
		\rowcolor{headerbg}
		\textbf{LLaVA-Video-7B+VISTA} &32& \textbf{66.6}~\textcolor{red}{(9.3\textuparrow~)} &  \textbf{63.2}~\textcolor{red}{(4.6\textuparrow~)} & \textbf{68.8}~\textcolor{red}{(0.9\textuparrow~)} & \textbf{54.4}~\textcolor{red}{(5.6\textuparrow~)} & \textbf{40.9}~\textcolor{red}{(6.5\textuparrow~)} \\
		\bottomrule[1.pt]
	\end{tabularx}
	\caption{Performance on video QA multiple choice benchmarks. \textbf{ES}, \textbf{MV}, \textbf{PT}, \textbf{VE}, and \textbf{VM} represent EgoSchema, MVBench, PerceptionTest, VideoEspresso, and VideoMMMU, respectively. * indicates the result we reproduced.}
	\label{tab:main results}
\end{table*}
\section{Experiment}
\subsection{Settings}
\textbf{Data and Evaluation.} We adopt five multiple choice video benchmarks that characterize complex video reasoning to highlight our performance including EgoSchema
\cite{DBLP:conf/nips/MangalamAM23}, PerceptionTest
\cite{patraucean2023perception}, 
VideoEspresso
\cite{han2025videoespresso}, 
MVBench
\cite{li2024mvbench} and VideoMMMU \cite{hu2025video}.
Advanced video abilities are required to address problems in these benchmarks, e.g. detailed comprehension, causal understanding, and contextual integration ability.

\textbf{Implementation Details.}
We separately use LLaVA-onevision-7B and LLaVA-Video-7B as our base models. These 7B models adopt Qwen2-7B-Instruct as LLM and use SigLIP \cite{zhai2023sigmoid} as image backbone.
Following LLaVA-Video, we represent each video as a sequence with maximum $T$ frames. 
Each frame is resized to 384x384 and represented by $M$ tokens.
$T$ and $M$ are individually initialized to 32 and 729 here.
Each frame is encoded via  SigLIP encoder and a two-layer MLP for projection.
Text and visual tokens are concatenated and fed into LLM.
We conducted all experiments on NVIDIA V100 32G GPUs.
\begin{table*}[t]
	\footnotesize
	\centering
	\setlength{\tabcolsep}{9.8pt}
	{
		\begin{tabular}{lcccc}
			\toprule[1.pt]
			\textbf{Model} & \textbf{EgoSchema}&\textbf{VideoEspresso}  &\textbf{MVBench} &\textbf{PerceptionTest}\\
			\midrule
			LLaVA-Video/OV &  57.3/60.1 & 48.8/44.0& 58.6/56.1& 67.9/57.1\\
			LLaVA-Video/OV + EVA& 61.4/63.8 & 50.7/44.5&59.1/56.0 & 68.0/60.3 \\
			LLaVA-Video/OV + DIR + EVA & 62.9/64.9 & 51.7/45.6 &60.4/57.1& 68.4/60.8 \\
			LLaVA-Video/OV + EVA + LRC& 65.8/66.8 & 53.7/47.3 & 61.3/58.3\ & 68.4/61.9 \\
			\rowcolor{headerbg}
			LLaVA-Video/OV + DIR + EVA + LRC & 66.6/67.8 & 54.4/47.9 &63.2/58.6& 68.8/62.4\\
			\bottomrule[1.pt]
		\end{tabular}
		\caption{Effectiveness of different modules.
			\textbf{DIR} means Dynamic Inference Routing. \textbf{EVA} means Explicit Visual Anchoring. \textbf{LRC} means Latent Reasoning Consensus.}
		\label{tab:effectiveness of different modules}
	}
\end{table*}

\subsection{Main Performance on Video QA}
In this section, we compare VISTA with the base
model, LLaVA-onevision and LLaVA-Video, on five commonly used video understanding benchmarks to prove validity of our training free framework.
In Table \ref{tab:main results}, we show main results of different video LLMs and our framework.
Across all evaluated benchmarks, the integration of VISTA framework has consistently achieved significant performance enhancements over existing state-of-the-art methods.
The comprehensive enhancement reveals two key findings:

\textbf{Consistent generalizability}. VISTA demonstrates universal compatibility with diverse base models,
achieving obvious gain over vanilla implementations.
Notably, when integrated with LLaVA-Video-7B, our framework attains \textbf{54.4\%} on VideoEspresso (vs. baseline 48.8\%) and \textbf{63.2\%} on MVBench (vs. baseline 58.6\%), outperforming 72B-parameter counterparts like LLaVA-OV-72B (59.4\%) and VideoLLaMA2-72B (62.0\%) without parameter expansion.

\begin{table}[t]
	\footnotesize
	\centering
	\newcolumntype{Y}{>{\centering\arraybackslash}X}
	\setlength{\tabcolsep}{3pt}
	
	\begin{tabularx}{\columnwidth}{l YYYY}
		\toprule[1.pt]
		\textbf{Model} & \textbf{ES} & \textbf{VE} & \textbf{PT} & \textbf{MV} \\
		\midrule
		\multicolumn{5}{l}{\emph{\textbf{Base: LLaVA-OneVision-7B}}} \\
		
		Naive prompting & 64.0 & 44.2 & 59.2 & 56.3 \\
		Majority voting & 64.6 & 44.9 & 59.8 & 57.0 \\
		Best of N       & 67.2 & 46.0 & 59.8 & 57.9 \\
		\rowcolor{headerbg}
		LRC (ours) & \textbf{67.8} & \textbf{47.9} & \textbf{62.4} & \textbf{58.6} \\
		\bottomrule[1.pt]
	\end{tabularx}
	
	\caption{Inference effects of different verification mechanisms. \textbf{ES}, \textbf{VE}, \textbf{PT}, and \textbf{MV} represent EgoSchema, VideoEspresso, PerceptionTest, and MVBench, respectively.}
	\label{tab:verification_methods}
\end{table}

\textbf{Task Specific Superiority}.
The framework shows particular strength in causal reasoning  and long-form understanding, validating its reasoning mechanism. 
However, the relatively narrow margin on perception-oriented tasks suggests greater challenges in low-level visual grounding.

These results confirm that our training-free framework effectively bridges the modality gap in video reasoning.
The performance highlights the great potential of our systematic reasoning framework over pure scale-based approaches.
\begin{table}[t]
	\footnotesize
	\centering
	\newcolumntype{Y}{>{\centering\arraybackslash}X}
	\setlength{\tabcolsep}{3pt}
	
	\begin{tabularx}{\columnwidth}{l YYYY}
		\toprule[1.pt]
		\textbf{Model Variant} & \textbf{ES} & \textbf{VE} & \textbf{PT} & \textbf{MV} \\
		\midrule
		\multicolumn{5}{l}{\emph{\textbf{Base: LLaVA-OneVison-7B}}} \\
		
		
		w/o Question & 63.8 & 46.8 & 58.5 & 56.2 \\
		\rowcolor{headerbg}
		w/ Question  & \textbf{64.4} & \textbf{47.9} & \textbf{60.3} & \textbf{58.3} \\
		w/o CoT      & 64.0 & 47.3 & 60.1 & 57.8 \\
		\rowcolor{headerbg}
		w/ CoT       & \textbf{64.4} & \textbf{47.9} & \textbf{60.3} & \textbf{58.3} \\
		\bottomrule[1.pt]
	\end{tabularx}
	
	\caption{Impact of standard CoT template and additional attention on input question. \textbf{ES}, \textbf{VE}, \textbf{PT}, and \textbf{MV} represent EgoSchema, VideoEspresso, PerceptionTest, and MVBench, respectively.}
	\label{tab:additional attention}
\end{table}

\subsection{Ablation Studies}
\textbf{Effectiveness of different modules.}
To further reveal the complex video reasoning mechanism, we explored the effectiveness of different modules in VISTA.
The experimental results are shown in Table \ref{tab:effectiveness of different modules}.
All modules in VISTA have robust effect boosts over different datasets,
with our proposed Latent Reasoning Consensus verification being the most prominent among them.
The results has a tangible dip with the Explicit Visual Anchoring phase removed. 

\textbf{Analysis on Verification Methods.}
We explored four different verification methods respectively, including naive prompting,
majority voting, best-of-N searching, and our proposed Latent Reasoning Consensus.
The experimental results of these four different methods are shown in Table \ref{tab:verification_methods}.
The experimental results further validate the effectiveness of our designed methodology.
It also reveals that mechanically applying CoT-related techniques may decrease reasoning performance, 
raising the need to design specifically according to task characteristics.

\textbf{Analysis on Additional Attention to the Question.}
Given the problem of hallucination that often occurs with existing models, we wonder if it would be better to guide the model to pay more attention on the content relevant to the question.
We explored a variety of prompting templates to accomplish this goal.
After sufficient experiments, We got an effective prompt to solve this problem: 
"Summarize the main content in the video, paying special attention to content related to the question: $Q$, unrelated part can be summarized more briefly."
By the way, the symbol $Q$ means the initial input question.
Results are shown in Table \ref{tab:additional attention}.

\textbf{Analysis on Validity of Standard CoT prompt.}
In order to further stimulate the model's reasoning ability, we tried to add the standard CoT prompt, “Let us think step by step”, in the Evidence-Grounded Deduction phase.
We attempted to analyze the impact on final results, shown in \ref{tab:additional attention}.
It can be inferred that the ultimate reasoning ability is mainly stimulated by our multi-level reasoning framework rather than the CoT prompt.

\begin{figure}[h]
	\centering
	\includegraphics[width=0.8\linewidth]{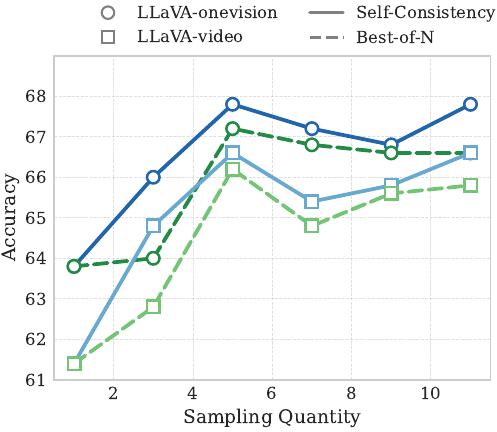}
	\caption{Trend of inference performance with number of samples}
	\label{fig:sampling quantity}
\end{figure}

\textbf{Analysis on Validity of Sample Size.}
In order to further analyze the trade-off between computational resources and the final performance, we experimented the effect of Latent Reasoning Consensus and best-of-N with different number of sampling. 
As shown in Figure \ref{fig:sampling quantity},
model performance increases steadily with the number of samples, saturating at around five samples.

\section{Related Work}

\textbf{Video Large Language Models.}
Recent Video-LLMs, represented by LLaVA-Video \cite{zhang2024video} and VideoLLaMA3 \cite{zhang2025videollama},
have made significant progress.
However, the reasoning mechanism lacks explicit logical modeling, and is essentially a shallow “perception-mapping” correlation.
Mainstream approaches typically employ SFT \cite{zhang2025video, wen2025agricot, muennighoff2025s, ye2025limoreasoning} or RL \cite{wang2025videorft, feng2025videor1reinforcingvideoreasoning, li2025videochat,jin2025videomem} to enhance reasoning reliability.
For instance, Video-R1 \cite{feng2025videor1reinforcingvideoreasoning} improves reasoning awareness by adding temporal constraints.
In contrast to these costly methods that often show limited improvements, our training-free VISTA achieves significant performance gains.

\textbf{Chain-of-Thought Reasoning.}
The CoT paradigm \cite{NEURIPS2022_9d560961} elicits reasoning by decomposing problems into steps. While widely adopted in NLP \cite{yao2023tree, Besta2023GraphGOT}, transferring CoT to video remains challenging.
Prior multimodal efforts typically fall into two categories:
(1) Training-Intensive methods (e.g., MM-CoT \cite{zhang2024multimodal}, LLaVA-CoT \cite{xu2024llavacot}, CoCoT \cite{zhang2024cocot}) that fine-tune models on structured data; and
(2) Tool-Dependent methods (e.g., VideoAgent \cite{fan2024videoagent}) that rely on external tools.
These approaches are limited by training overhead or fixed pipelines. VISTA achieves this via a flexible, training-free mechanism.

\textbf{Test Time Scaling.}
Complex reasoning framework consists of System 1 (fast reactive decision-making) and System 2 (slow hierarchical reasoning) \cite{wang2025multimodalchainofthoughtreasoningcomprehensive}.
With OpenAI o1 \cite{jaech2024openai} and Deepseek R1 \cite{deepseekai2025deepseekr1incentivizingreasoningcapability}, Test-Time Scaling has attracted widespread attention.
It shows unique advantages by dynamically fusing the intuitive prior of System 1 with the slow inference mechanism of System 2.
The core innovation lies in the optimization of the reasoning strategy in the testing phase: it realizes the hierarchical solution by dynamically adjusting the reasoning steps.
This hybrid architecture of “intuition-guided, logic-verified” provides a viable direction for video understanding research.

\section{Conclusion}
In this paper, we identify Semantic Inertia as a primary bottleneck in Video-LLMs, where language priors suppress valid visual evidence during complex reasoning. To address this, we propose VISTA, a training-free framework that grounds generation in explicit visual facts through dynamic routing and structured anchoring. Our extensive experiments demonstrate that VISTA effectively mitigates hallucinations and unlocks perceptual capabilities, achieving performance competitive with state-of-the-art proprietary models. These results validate the efficacy of inference-time scaling strategies, offering a scalable path toward robust System 2 video reasoning.

\section*{Limitations} While VISTA demonstrates significant improvements in mitigating semantic inertia and enhancing complex video reasoning, it presents several limitations. Firstly, the framework incurs increased computational overhead and inference latency due to the additional token generation required for Explicit Visual Anchoring and the multi-path sampling strategy in Latent Reasoning Consensus,
which may restrict its deployment in real-time or resource-constrained scenarios. Furthermore, as a training-free framework, VISTA's upper bound is inherently constrained by the base model's capabilities, and any hallucinations occurring during the intermediate visual evidence materialization phase can propagate errors into the final deduction.

\bibliography{main}

@inproceedings{wang2024internvideo2,
  title={Internvideo2: Scaling foundation models for multimodal video understanding},
  author={Wang, Yi and Li, Kunchang and Li, Xinhao and Yu, Jiashuo and He, Yinan and Chen, Guo and Pei, Baoqi and Zheng, Rongkun and Wang, Zun and Shi, Yansong and others},
  booktitle={European Conference on Computer Vision},
  pages={396--416},
  year={2024},
  organization={Springer}
}

@inproceedings{ren2024timechat,
  title={Timechat: A time-sensitive multimodal large language model for long video understanding},
  author={Ren, Shuhuai and Yao, Linli and Li, Shicheng and Sun, Xu and Hou, Lu},
  booktitle={Proceedings of the IEEE/CVF Conference on Computer Vision and Pattern Recognition},
  pages={14313--14323},
  year={2024}
}

@article{wen2025agricot,
  title={AgriCoT: A Chain-of-Thought Benchmark for Evaluating Reasoning in Vision-Language Models for Agriculture},
  author={Wen, Yibin and Li, Qingmei and Ye, Zi and Zhang, Jiarui and Wu, Jing and Mai, Zurong and Lou, Shuohong and Chen, Yuhang and Huang, Henglian and Fan, Xiaoya and others},
  journal={arXiv preprint arXiv:2511.23253},
  year={2025}
}

@article{wang2025videorft,
  title={VideoRFT: Incentivizing Video Reasoning Capability in MLLMs via Reinforced Fine-Tuning},
  author={Wang, Qi and Yu, Yanrui and Yuan, Ye and Mao, Rui and Zhou, Tianfei},
  journal={arXiv preprint arXiv:2505.12434},
  year={2025}
}

@article{zhang2025video,
  title={Video-CoT: A Comprehensive Dataset for Spatiotemporal Understanding of Videos Based on Chain-of-Thought},
  author={Zhang, Shuyi and Hao, Xiaoshuai and Tang, Yingbo and Zhang, Lingfeng and Wang, Pengwei and Wang, Zhongyuan and Ma, Hongxuan and Zhang, Shanghang},
  journal={arXiv preprint arXiv:2506.08817},
  year={2025}
}

@inproceedings{DBLP:conf/nips/MangalamAM23,
  author={Karttikeya Mangalam and Raiymbek Akshulakov and Jitendra Malik},
  title={EgoSchema: A Diagnostic Benchmark for Very Long-form Video Language Understanding},
  year={2023},
  cdate={1672531200000},
  url={http://papers.nips.cc/paper_files/paper/2023/hash/90ce332aff156b910b002ce4e6880dec-Abstract-Datasets_and_Benchmarks.html},
  booktitle={NeurIPS}
}

@article{yao2023tree,
  title={Tree of thoughts: Deliberate problem solving with large language models},
  author={Yao, Shunyu and Yu, Dian and Zhao, Jeffrey and Shafran, Izhak and Griffiths, Tom and Cao, Yuan and Narasimhan, Karthik},
  journal={Advances in neural information processing systems},
  volume={36},
  pages={11809--11822},
  year={2023}
}

@inproceedings{han2025videoespresso,
  title={Videoespresso: A large-scale chain-of-thought dataset for fine-grained video reasoning via core frame selection},
  author={Han, Songhao and Huang, Wei and Shi, Hairong and Zhuo, Le and Su, Xiu and Zhang, Shifeng and Zhou, Xu and Qi, Xiaojuan and Liao, Yue and Liu, Si},
  booktitle={Proceedings of the Computer Vision and Pattern Recognition Conference},
  pages={26181--26191},
  year={2025}
}

@inproceedings{NEURIPS2022_9d560961,
 author = {Wei, Jason and Wang, Xuezhi and Schuurmans, Dale and Bosma, Maarten and ichter, brian and Xia, Fei and Chi, Ed and Le, Quoc V and Zhou, Denny},
 booktitle = {Advances in Neural Information Processing Systems},
 editor = {S. Koyejo and S. Mohamed and A. Agarwal and D. Belgrave and K. Cho and A. Oh},
 pages = {24824--24837},
 publisher = {Curran Associates, Inc.},
 title = {Chain-of-Thought Prompting Elicits Reasoning in Large Language Models},
 url = {https://proceedings.neurips.cc/paper_files/paper/2022/file/9d5609613524ecf4f15af0f7b31abca4-Paper-Conference.pdf},
 volume = {35},
 year = {2022}
}

@misc{deepseekai2025deepseekr1incentivizingreasoningcapability,
      title={DeepSeek-R1: Incentivizing Reasoning Capability in LLMs via Reinforcement Learning}, 
      author={DeepSeek-AI},
      year={2025},
      eprint={2501.12948},
      archivePrefix={arXiv},
      primaryClass={cs.CL},
      url={https://arxiv.org/abs/2501.12948}, 
}

@article{damonlpsg2024videollama2,
  title={VideoLLaMA 2: Advancing Spatial-Temporal Modeling and Audio Understanding in Video-LLMs},
  author={Cheng, Zesen and Leng, Sicong and Zhang, Hang and Xin, Yifei and Li, Xin and Chen, Guanzheng and Zhu, Yongxin and Zhang, Wenqi and Luo, Ziyang and Zhao, Deli and Bing, Lidong},
  journal={arXiv preprint arXiv:2406.07476},
  year={2024},
  url = {https://arxiv.org/abs/2406.07476}
}

@misc{wang2025multimodalchainofthoughtreasoningcomprehensive,
      title={Multimodal Chain-of-Thought Reasoning: A Comprehensive Survey}, 
      author={Yaoting Wang and Shengqiong Wu and Yuecheng Zhang and William Wang and Ziwei Liu and Jiebo Luo and Hao Fei},
      year={2025},
      eprint={2503.12605},
      archivePrefix={arXiv},
      primaryClass={cs.CV},
      url={https://arxiv.org/abs/2503.12605}, 
}

@article{jaech2024openai,
  title={Openai o1 system card},
  author={Jaech, Aaron and Kalai, Adam and Lerer, Adam and Richardson, Adam and El-Kishky, Ahmed and Low, Aiden and Helyar, Alec and Madry, Aleksander and Beutel, Alex and Carney, Alex and others},
  journal={arXiv preprint arXiv:2412.16720},
  year={2024}
}

@article{wu2024longvideobench,
  title={Longvideobench: A benchmark for long-context interleaved video-language understanding},
  author={Wu, Haoning and Li, Dongxu and Chen, Bei and Li, Junnan},
  journal={Advances in Neural Information Processing Systems},
  volume={37},
  pages={28828--28857},
  year={2024}
}

@inproceedings{li2024mvbench,
  title={Mvbench: A comprehensive multi-modal video understanding benchmark},
  author={Li, Kunchang and Wang, Yali and He, Yinan and Li, Yizhuo and Wang, Yi and Liu, Yi and Wang, Zun and Xu, Jilan and Chen, Guo and Luo, Ping and others},
  booktitle={Proceedings of the IEEE/CVF Conference on Computer Vision and Pattern Recognition},
  pages={22195--22206},
  year={2024}
}

@article{achiam2023gpt,
  title={Gpt-4 technical report},
  author={Achiam, Josh and Adler, Steven and Agarwal, Sandhini and Ahmad, Lama and Akkaya, Ilge and Aleman, Florencia Leoni and Almeida, Diogo and Altenschmidt, Janko and Altman, Sam and Anadkat, Shyamal and others},
  journal={arXiv preprint arXiv:2303.08774},
  year={2023}
}

@article{team2023gemini,
  title={Gemini: a family of highly capable multimodal models},
  author={Team, Gemini and Anil, Rohan and Borgeaud, Sebastian and Alayrac, Jean-Baptiste and Yu, Jiahui and Soricut, Radu and Schalkwyk, Johan and Dai, Andrew M and Hauth, Anja and Millican, Katie and others},
  journal={arXiv preprint arXiv:2312.11805},
  year={2023}
}

@misc{google_gemini_1.5_2024,
  author = {{Google}},
  title = {Introducing Gemini 1.5, {Google}’s next-generation {AI} model},
  url = {https://blog.google/technology/ai/google-gemini-next-generationmodel-february-2024/},
  year = {2024},
  note = {Accessed: 2024-06-10}
}

@misc{openai_gpt4o_2024,
  author = {{OpenAI}},
  title = {Hello {GPT-4o}},
  url = {https://openai.com/index/hello-gpt-4o/},
  year = {2024},
  month = {May}
}

@article{qu2024ts,
  title={TS-LLaVA: Constructing Visual Tokens through Thumbnail-and-Sampling for Training-Free Video Large Language Models},
  author={Qu, Tingyu and Li, Mingxiao and Tuytelaars, Tinne and Moens, Marie-Francine},
  journal={arXiv preprint arXiv:2411.11066},
  year={2024}
}

@inproceedings{liu2025nvila,
  title={Nvila: Efficient frontier visual language models},
  author={Liu, Zhijian and Zhu, Ligeng and Shi, Baifeng and Zhang, Zhuoyang and Lou, Yuming and Yang, Shang and Xi, Haocheng and Cao, Shiyi and Gu, Yuxian and Li, Dacheng and others},
  booktitle={Proceedings of the Computer Vision and Pattern Recognition Conference},
  pages={4122--4134},
  year={2025}
}

@article{zhang2024long,
  title={Long context transfer from language to vision},
  author={Zhang, Peiyuan and Zhang, Kaichen and Li, Bo and Zeng, Guangtao and Yang, Jingkang and Zhang, Yuanhan and Wang, Ziyue and Tan, Haoran and Li, Chunyuan and Liu, Ziwei},
  journal={arXiv preprint arXiv:2406.16852},
  year={2024}
}

@article{zhang2024internlm,
  title={Internlm-xcomposer-2.5: A versatile large vision language model supporting long-contextual input and output},
  author={Zhang, Pan and Dong, Xiaoyi and Zang, Yuhang and Cao, Yuhang and Qian, Rui and Chen, Lin and Guo, Qipeng and Duan, Haodong and Wang, Bin and Ouyang, Linke and others},
  journal={arXiv preprint arXiv:2407.03320},
  year={2024}
}

@article{li2024llava,
  title={Llava-onevision: Easy visual task transfer},
  author={Li, Bo and Zhang, Yuanhan and Guo, Dong and Zhang, Renrui and Li, Feng and Zhang, Hao and Zhang, Kaichen and Zhang, Peiyuan and Li, Yanwei and Liu, Ziwei and others},
  journal={arXiv preprint arXiv:2408.03326},
  year={2024}
}

@article{zhang2024video,
  title={Video instruction tuning with synthetic data},
  author={Zhang, Yuanhan and Wu, Jinming and Li, Wei and Li, Bo and Ma, Zejun and Liu, Ziwei and Li, Chunyuan},
  journal={arXiv preprint arXiv:2410.02713},
  year={2024}
}

@article{zhang2025videollama,
  title={VideoLLaMA 3: Frontier Multimodal Foundation Models for Image and Video Understanding},
  author={Zhang, Boqiang and Li, Kehan and Cheng, Zesen and Hu, Zhiqiang and Yuan, Yuqian and Chen, Guanzheng and Leng, Sicong and Jiang, Yuming and Zhang, Hang and Li, Xin and others},
  journal={arXiv preprint arXiv:2501.13106},
  year={2025}
}

@misc{feng2025videor1reinforcingvideoreasoning,
      title={Video-R1: Reinforcing Video Reasoning in MLLMs}, 
      author={Kaituo Feng and Kaixiong Gong and Bohao Li and Zonghao Guo and Yibing Wang and Tianshuo Peng and Benyou Wang and Xiangyu Yue},
      year={2025},
      eprint={2503.21776},
      archivePrefix={arXiv},
      primaryClass={cs.CV},
      url={https://arxiv.org/abs/2503.21776}, 
}

@article{li2025system,
  title={From system 1 to system 2: A survey of reasoning large language models},
  author={Li, Zhong-Zhi and Zhang, Duzhen and Zhang, Ming-Liang and Zhang, Jiaxin and Liu, Zengyan and Yao, Yuxuan and Xu, Haotian and Zheng, Junhao and Wang, Pei-Jie and Chen, Xiuyi and others},
  journal={arXiv preprint arXiv:2502.17419},
  year={2025}
}

@inproceedings{
muennighoff2025s,
title={s1: Simple test-time scaling},
author={Niklas Muennighoff and Zitong Yang and Weijia Shi and Xiang Lisa Li and Li Fei-Fei and Hannaneh Hajishirzi and Luke Zettlemoyer and Percy Liang and Emmanuel Candes and Tatsunori Hashimoto},
booktitle={Workshop on Reasoning and Planning for Large Language Models},
year={2025},
url={https://openreview.net/forum?id=LdH0vrgAHm}
}

@misc{ye2025limoreasoning,
      title={LIMO: Less is More for Reasoning}, 
      author={Yixin Ye and Zhen Huang and Yang Xiao and Ethan Chern and Shijie Xia and Pengfei Liu},
      year={2025},
      eprint={2502.03387},
      archivePrefix={arXiv},
      primaryClass={cs.CL},
      url={https://arxiv.org/abs/2502.03387}, 
}

@inproceedings{Besta2023GraphGOT,
    title={Graph of Thoughts: Solving Elaborate Problems with Large Language Models},
    author={Maciej Besta and Nils Blach and Aleš Kubíček and Robert Gerstenberger and Lukas Gianinazzi and Joanna Gajda and Tomasz Lehmann and Michal Podstawski and H. Niewiadomski and P. Nyczyk and Torsten Hoefler},
    year={2023},
    url={https://www.semanticscholar.org/paper/aade40af0d85b0b4fe15c97f6222d5c2e4d6d9b3},
    booktitle={AAAI Conference on Artificial Intelligence},
}

@misc{xu2024llavacot,
      title={LLaVA-CoT: Let Vision Language Models Reason Step-by-Step},
      author={Guowei Xu and Peng Jin and Hao Li and Yibing Song and Lichao Sun and Li Yuan},
      year={2024},
      eprint={2411.10440},
      archivePrefix={arXiv},
      primaryClass={cs.CV},
      url={https://arxiv.org/abs/2411.10440},
}

@article{
zhang2024multimodal,
title={Multimodal Chain-of-Thought Reasoning in Language Models},
author={Zhuosheng Zhang and Aston Zhang and Mu Li and hai zhao and George Karypis and Alex Smola},
journal={Transactions on Machine Learning Research},
issn={2835-8856},
year={2024},
url={https://openreview.net/forum?id=y1pPWFVfvR},
note={}
}

@inproceedings{patraucean2023perception,
      title={Perception Test: A Diagnostic Benchmark for Multimodal Video Models}, 
      author={Viorica Pătrăucean and Lucas Smaira and Ankush Gupta and Adrià Recasens Continente and Larisa Markeeva and Dylan Banarse and Skanda Koppula and Joseph Heyward and Mateusz Malinowski and Yi Yang and Carl Doersch and Tatiana Matejovicova and Yury Sulsky and Antoine Miech and Alex Frechette and Hanna Klimczak and Raphael Koster and Junlin Zhang and Stephanie Winkler and Yusuf Aytar and Simon Osindero and Dima Damen and Andrew Zisserman and João Carreira},
      booktitle={Advances in Neural Information Processing Systems},
      year={2023},
      url={https://openreview.net/forum?id=HYEGXFnPoq}
}

@inproceedings{zhai2023sigmoid,
  title={Sigmoid loss for language image pre-training},
  author={Zhai, Xiaohua and Mustafa, Basil and Kolesnikov, Alexander and Beyer, Lucas},
  booktitle={Proceedings of the IEEE/CVF international conference on computer vision},
  pages={11975--11986},
  year={2023}
}

@misc{xu2024pllava,
      title={PLLaVA : Parameter-free LLaVA Extension from Images to Videos for Video Dense Captioning}, 
      author={Lin Xu and Yilin Zhao and Daquan Zhou and Zhijie Lin and See Kiong Ng and Jiashi Feng},
      year={2024},
      eprint={2404.16994},
      archivePrefix={arXiv},
      primaryClass={cs.CV}
}

@article{hu2025video,
  title={Video-MMMU: Evaluating Knowledge Acquisition from Multi-Discipline Professional Videos},
  author={Hu, Kairui and Wu, Penghao and Pu, Fanyi and Xiao, Wang and Zhang, Yuanhan and Yue, Xiang and Li, Bo and Liu, Ziwei},
  journal={arXiv preprint arXiv:2501.13826},
  year={2025}
}

@article{li2025videochat,
  title={VideoChat-R1: Enhancing Spatio-Temporal Perception via Reinforcement Fine-Tuning},
  author={Li, Xinhao and Yan, Ziang and Meng, Desen and Dong, Lu and Zeng, Xiangyu and He, Yinan and Wang, Yali and Qiao, Yu and Wang, Yi and Wang, Limin},
  journal={arXiv preprint arXiv:2504.06958},
  year={2025}
}

@article{zhang2024cocot,
  title={Cocot: Contrastive chain-of-thought prompting for large multimodal models with multiple image inputs},
  author={Zhang, Daoan and Yang, Junming and Lyu, Hanjia and Jin, Zijian and Yao, Yuan and Chen, Mingkai and Luo, Jiebo},
  journal={arXiv preprint arXiv:2401.02582},
  year={2024}
}

@inproceedings{fan2024videoagent,
  title={Videoagent: A memory-augmented multimodal agent for video understanding},
  author={Fan, Yue and Ma, Xiaojian and Wu, Rujie and Du, Yuntao and Li, Jiaqi and Gao, Zhi and Li, Qing},
  booktitle={European Conference on Computer Vision},
  pages={75--92},
  year={2024},
  organization={Springer}
}

@inproceedings{
zhang2025llavamini,
title={{LL}a{VA}-Mini: Efficient Image and Video Large Multimodal Models with One Vision Token},
author={Shaolei Zhang and Qingkai Fang and Zhe Yang and Yang Feng},
booktitle={The Thirteenth International Conference on Learning Representations},
year={2025},
url={https://openreview.net/forum?id=UQJ7CDW8nb}
}

@article{jin2025videomem,
  title={VideoMem: Enhancing Ultra-Long Video Understanding via Adaptive Memory Management},
  author={Jin, Hongbo and Wang, Qingyuan and Zhang, Wenhao and Liu, Yang and Cheng, Sijie},
  journal={arXiv preprint arXiv:2512.04540},
  year={2025}
}

@inproceedings{
li2025llavanextinterleave,
title={{LL}a{VA}-Ne{XT}-Interleave: Tackling Multi-image, Video, and 3D in Large Multimodal Models},
author={Feng Li and Renrui Zhang and Hao Zhang and Yuanhan Zhang and Bo Li and Wei Li and Zejun MA and Chunyuan Li},
booktitle={The Thirteenth International Conference on Learning Representations},
year={2025},
url={https://openreview.net/forum?id=oSQiao9GqB}
}

@inproceedings{li2024llamavid,
  title={LLaMA-VID: An Image is Worth 2 Tokens in Large Language Models},
  author={Li, Yanwei and Wang, Chengyao and Jia, Jiaya},
  journal={European Conference on Computer Vision},
  year={2024}
}

@article{hurst2024gpt,
  title={Gpt-4o system card},
  author={Hurst, Aaron and Lerer, Adam and Goucher, Adam P and Perelman, Adam and Ramesh, Aditya and Clark, Aidan and Ostrow, AJ and Welihinda, Akila and Hayes, Alan and Radford, Alec and others},
  journal={arXiv preprint arXiv:2410.21276},
  year={2024}
}

\clearpage
\appendix
\appendix


\section{Case Study}
\subsection{Limitations observed on other datasets}

Although the \textbf{VISTA} framework improves significantly in model complex reasoning, limitations have been observed for simple perception-based tasks.

\begin{tcolorbox}[
	colframe=black,
	colback=gray!10,
	sharp corners=southwest,
	enhanced,
	breakable,
	listing only,
	listing options={
		basicstyle=\ttfamily\small,
		breaklines=true,
	}
	]
	\textbf{Dataset:} \textit{VideoMME}
	
	\textbf{Video:} \textit{24i4ncHuf6A}
	
	\textbf{Question:} \textit{According to the video, how many individuals were in the car when Archduke Franz Ferdinand was assassinated?}
	
	\textbf{Answer:} \textit{A. Three}
	
	\textbf{Candidates:}
	\begin{itemize}[noitemsep,topsep=0pt,parsep=0pt,partopsep=0pt]
		\item \textit{A. Three}
		\item \textit{B. Two}
		\item \textit{C. One}
		\item \textit{D. Four}
	\end{itemize}
\end{tcolorbox}

\noindent\textbf{Issue:} This question focuses on a specific detail at a particular moment in the video. This type of problem relies more on model perception and modal alignment capabilities. In such contexts, the reasoning capability of the \textbf{VISTA} framework does not function effectively.

\begin{tcolorbox}[
	colframe=black,
	colback=gray!10,
	sharp corners=southwest,
	enhanced,
	breakable,
	listing only,
	listing options={
		basicstyle=\ttfamily\small,
		breaklines=true,
	}
	]
	\textbf{Dataset:} \textit{VideoMME}
	
	\textbf{Video:} \textit{LCtOpCi5r2s}
	
	\textbf{Question:} \textit{Which item was not featured in the video?}
	
	\textbf{Answer:} \textit{A. Three}
	
	\textbf{Candidates:}
	\begin{itemize}[noitemsep,topsep=0pt,parsep=0pt,partopsep=0pt]
		\item \textit{A. Balance scale}
		\item \textit{B. Traffic light}
		\item \textit{C. Gavel}
		\item \textit{D. Magnifying glass}
	\end{itemize}
\end{tcolorbox}

\noindent\textbf{Issue:} This video scene is relatively complex and diverse, and the question is focused on perceiving a particular object or feature. In this case, the enhancement of reasoning ability is not enough to compensate for the lack of perception ability, and \textbf{VISTA} is more suitable for scenarios requiring logical deduction rather than pure visual search.

\subsection{A Typical Case Comparison}
\begin{figure*}[t!]
	\centering
	\includegraphics[width=\linewidth]{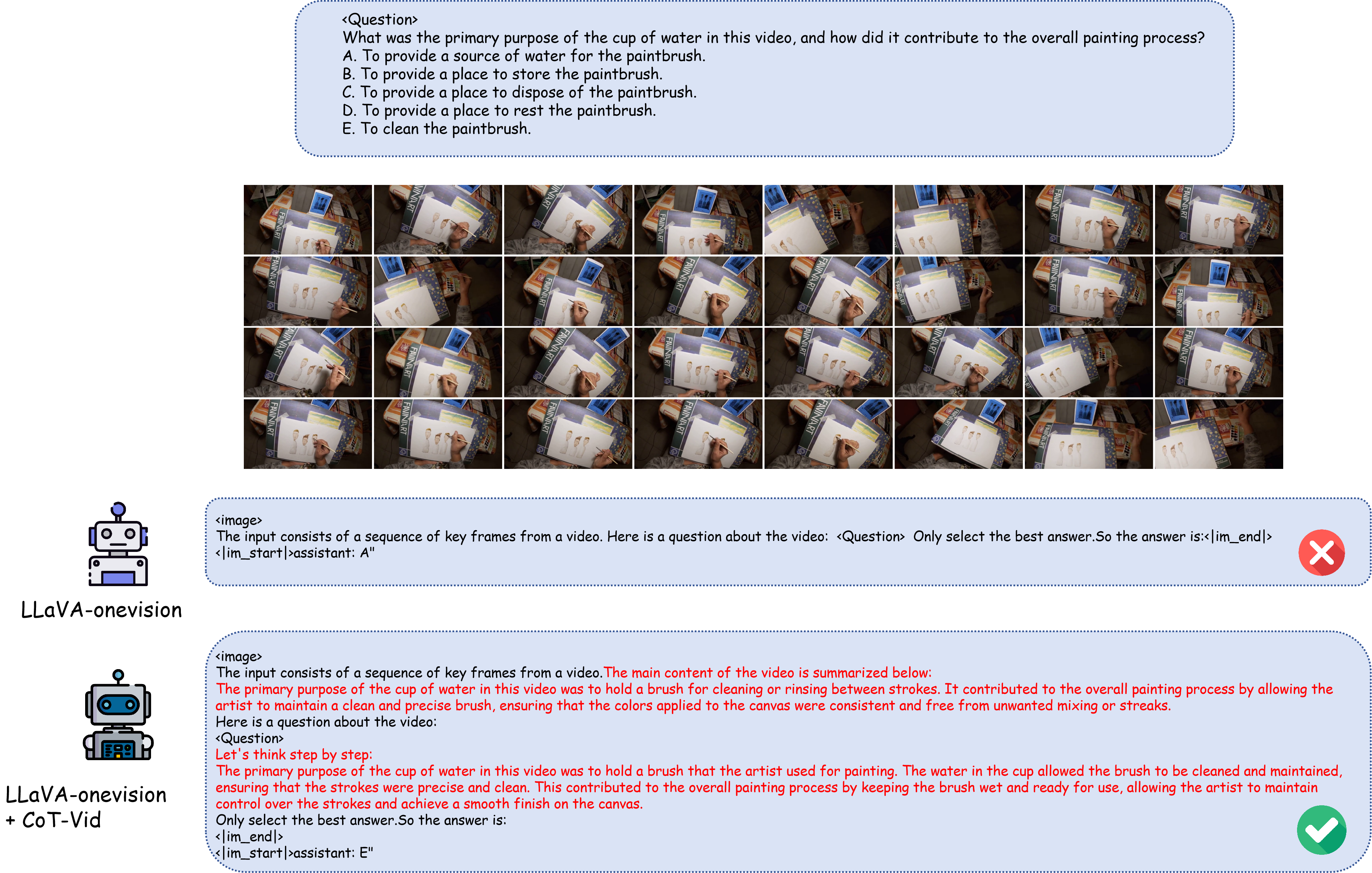}
	\caption{A typical case to illustrate the superiority of \textbf{VISTA}.}
	\label{fig:case study}
\end{figure*}

We show a typical case comparison result to demonstrate the effectiveness of the \textbf{VISTA} framework in Figure \ref{fig:case study}.
In this specific example, the question is: \textit{"What was the primary purpose of the cup of water in this video, and how did it contribute to the overall painting process?"}
The challenge of this question lies in that it has multiple subquestions and requires a comprehensive understanding of the video as a whole.
Moreover, the video frames are highly similar to each other, which increases the need for the model to focus on the dynamics at the details.
It is easy to observe that \textbf{VISTA} makes the model's reasoning process more interpretable and gains better inference performance.

\section{Details of Pilot Experiment}

In this section, we provide a comprehensive description of the data curation and annotation process for the pilot study. We specifically selected 100 hard negative samples from the MVBench validation set where the base model failed to predict the correct option. Our primary goal was to isolate the atomic visual facts required to answer these complex queries by decomposing the reasoning process.

\subsection{Atomic Visual Fact Generation}
To extract the key visual evidence, we utilized a multimodal expert model (Qwen-VL-Max) prompted to deconstruct the reasoning process into a chain of atomic visual facts. Unlike simple question generation, the prompt was strictly designed to break down the logic into sequential visual steps (Visual Premises), ensuring that the collection of facts is sufficient to deduce the answer without seeing the video. The specific prompt template used is as follows:

\begin{tcolorbox}[
	colframe=black,
	colback=gray!10,
	coltitle=white,
	colbacktitle=black,
	fonttitle=\bfseries,
	title={Atomic Visual Fact Extraction Prompt (1/2): Context \& Goal},
	sharp corners=southwest,
	enhanced,
	center title,
	boxrule=1pt
	]
	\small
	\textbf{Role:} You are a Lead Visual Forensic Analyst.
	
	\textbf{Task:} Deconstruct a complex video reasoning problem into a chain of ATOMIC visual facts.
	
	\textbf{Input:}
	\begin{itemize}[nosep, leftmargin=*]
		\item User Question: "[Input Question]"
		\item Correct Answer: "[Ground Truth]"
		\item Context: A smaller model FAILED to answer this because it missed visual details.
	\end{itemize}
	
	\textbf{Goal:}
	You must identify a SET of 3-5 atomic visual facts. 
	\textit{Crucial Requirement:} If a blind person reads ONLY your list of visual facts, they MUST be able to logically deduce the "Correct Answer" without seeing the video.
\end{tcolorbox}

\begin{tcolorbox}[
	colframe=black,
	colback=gray!10,
	coltitle=white,
	colbacktitle=black,
	fonttitle=\bfseries,
	title={Atomic Visual Fact Extraction Prompt (2/2): Instructions \& Format},
	sharp corners=southwest,
	enhanced,
	center title,
	boxrule=1pt
	]
	\small
	\textbf{Step-by-Step Instructions:}
	\begin{enumerate}[nosep, leftmargin=*]
		\item Analyze the logic required to go from the Question to the Correct Answer.
		\item Break this logic down into sequential visual steps (Visual Premises).
		\item For each step, create:
		\begin{itemize}[nosep]
			\item A "Visual Fact": A declarative statement of what is seen (e.g., "The traffic light is red").
			\item A "Binary Probe": A simple YES/NO question to check this fact (e.g., "Is the traffic light red?").
			\item The "Answer": "Yes" or "No".
		\end{itemize}
	\end{enumerate}
	
	\textbf{Constraints:}
	\begin{itemize}[nosep, leftmargin=*]
		\item Probe questions must be VISUAL and BASIC (perception level).
		\item Avoid high-level reasoning in the probes (e.g., don't ask "Is he angry?", ask "Is he frowning?").
		\item The collection of facts must be SUFFICIENT to support the final answer.
	\end{itemize}
	
	\textbf{Output Format (JSON Only):}
	\begin{verbatim}
		{
			"reasoning_chain": [
			{
				"step_id": 1,
				"visual_fact": "...", 
				"binary_probe": "...",
				"probe_answer": "Yes/No"
			},
			...
			],
			"sufficiency_check": "Explain why..."
		}
	\end{verbatim}
\end{tcolorbox}

\subsection{Human Verification and Statistics}
To ensure a high-quality benchmark, we validated the decomposed reasoning chains through a rigorous human-in-the-loop process involving three distinct annotators.

\textbf{Annotation Protocol.} The annotators reviewed each generated reasoning chain against the original video. They assessed validity based on three criteria: (1) \textbf{Atomicity}, ensuring each probe asks about a basic perceptual detail rather than high-level semantics; (2) \textbf{Factual Correctness}, ensuring the ground truth for each probe is objectively correct; and (3) \textbf{Sufficiency}, ensuring the sequence of facts logically supports the final answer. Any ambiguity was flagged and adjudicated by a senior annotator.

\textbf{Dataset Statistics.} 
Following this rigorous screening and quality control process, we retained only the validated reasoning chains. From the initial pool of 100 failure cases, we ultimately curated a final dataset comprising 416 atomic visual probe questions. 
The distribution of the reasoning chain lengths is as follows: the minimum length is 3 steps, and the maximum is 6 steps. The majority of samples (75\%) required exactly 4 steps, while 19\% required 5 steps, and the remaining covered 3 or 6 steps.

\begin{table*}[t!] 
	\footnotesize
	\centering
	\resizebox{\textwidth}{!}{ 
		\begin{tabular}{lccccc}
			\toprule[1.pt]
			\textbf{\diagbox[]{Model(7B)}{EgoSchema}} & \textbf{fact retrieval}&\textbf{process description}&\textbf{causal reasoning}&\textbf{theme summary}&\textbf{other} \\
			\midrule
			LLaVA-onevision &58.1&50.0&66.0&55.8&65.0\\
			\rowcolor{gray!10}
			LLaVA-onevision+\textbf{VISTA}&67.7~\textcolor{red}{(9.6\textuparrow~)}&55.0~\textcolor{red}{(5.0\textuparrow~)} &73.0~\textcolor{red}{(7.0\textuparrow~)}&57.7~\textcolor{red}{(1.9\textuparrow~)}&71.5~\textcolor{red}{(6.5\textuparrow~)}\\
			LLaVA-Video & 53.2&52.5&69.2&55.8&62.0\\
			\rowcolor{gray!10}
			LLaVA-Video+\textbf{VISTA}&59.7~\textcolor{red}{(6.5\textuparrow~)}&55.0~\textcolor{red}{(2.5\textuparrow~)}&71.7~\textcolor{red}{(2.5\textuparrow~)}&59.6~\textcolor{red}{(3.8\textuparrow~)}&71.5~\textcolor{red}{(9.5\textuparrow~)}\\
			\bottomrule[1.pt]
		\end{tabular}
	}
	\caption{Performance on typical subquestions.}
	\label{tab:subquestions}
\end{table*}
\section{Details of Dynamic Inference Routing}
\subsection{Full Details of Question Feature Table}\label{full details of feature table}

\begin{tcolorbox}[
	title=Question Features Table,
	colback=gray!10, 
	colframe=black!80,
	fonttitle=\bfseries,
	sharp corners,
	boxrule=0.5pt,
	top=3mm,
	bottom=3mm,
	]
	\begin{itemize}[
		leftmargin=*,
		nosep,
		itemsep=3mm,
		labelsep=1ex,
		before=\raggedright
		]
		\item $\square$ {Fact Retrieval}: 
		\textcolor{gray!70}{
			how many,   
			name the,
			identify the,
			key tools,
			specific item,
			which material
		}
		\item $\square$ {Process Description}: 
		\textcolor{gray!70}{
			describe the process,
			steps taken,
			sequence of actions,
			from start to finish,
			progress,
			workflow,
			procedures,
			step-by-step,
			sequentially
		}
		\item $\blacksquare$ {Causal Reasoning}: 
		\textcolor{gray!70}{
			explain,
			infer,
			deduce,
			why,
			how did,
			contribute to,
			result in,
			because,
			rationale behind,
			led to,
			impact of,
			relationship between
		}
		\item $\blacksquare$ {Theme Summary}:  
		\textcolor{gray!70}{
			overarching theme,
			primary objective,
			main goal,
			central purpose,
			fundamental intention,
			core focus,
			essential aim,
			principal motivation,
			underlying narrative
		}
		\item $\blacksquare$ {Comparative Analysis}:  
		\textcolor{gray!70}{
			compare,
			contrast,
			similarities,
			differences,
			distinguish from,
			relative importance,
			more significant,
			versus,
			whereas,
			unlike
		}
		\item $\blacksquare$ {Behavior Inference}:  
		\textcolor{gray!70}{
			infer,
			deduce,
			possible reason,
			underlying motivation,
			significance of,
			implications,
			hidden purpose,
			unspoken intention,
			symbolic meaning
		}
		\item $\square$ {Key Moment}:
		\textcolor{gray!70}{
			critical step,
			turning point,
			pivotal moment,
			decisive action,
			crucial stage,
			defining event,
			watershed moment,
			game-changing
		}
		\item $\blacksquare$ {Interaction Analysis}:
		\textcolor{gray!70}{
			interaction between,
			collaboration,
			communication,
			dynamic with,
			relationship with,
			coordination,
			exchange with,
			interplay,
			cooperation,
			conflict
		}
		\item $\square$ {Others}
	\end{itemize}
	
	\vspace{2mm}
	\hfill
	\begin{tabular}{@{}l@{}}
		$\blacksquare$ w/ multistep reasoning (VISTA) \\
		$\square$ w/o multistep reasoning (Direct Inference)
	\end{tabular}
\end{tcolorbox}

\subsection{Performance on Subquestions}
Table \ref{tab:subquestions} shows the performance improvement of the \textbf{VISTA} framework on each of our predefined typical subquestions.

\subsection{Alternative Routing Mechanism: Question Assessment Pipeline}\label{another question pipeline}

We propose an alternative question assessment pipeline that combines syntactic and lexical analysis through three key dimensions, aggregated via a weighted scoring mechanism. 
The input question first goes through two branches, syntactic analysis and lexical analysis respectively.
Then syntactic and lexical features are fused together and computed to get a complexity score.
This complexity score is used to determine if a question of the current difficulty requires complex reasoning (i.e., routing to \textbf{VISTA}).

\textbf{Syntactic Complexity Analysis}.
The syntactic analysis module is implemented through two core metrics:
dependency count and clause count.

\textbf{Dependency relations count} captures surface-level complexity through token enumeration, shown in Equation \ref{dependency},
where $\mathcal{G}$ represents the dependency graph.

\begin{equation}
	N_{dep} = \sum_{(h \to d) \in \mathcal{G}} \mathbb{I}(h.pos \neq d.pos )
	\label{dependency}
\end{equation}

\textbf{Clause detection mechanism} identifies subordinate clauses through \texttt{mark} dependencies, shown in Equation \ref{clause},
where $\mathcal{T}$ denotes the parsed tokens.

\begin{equation}
	N_{clause} = \sum_{t \in \mathcal{T}} \delta(t.dep = \texttt{"mark"})
	\label{clause}
\end{equation}

This targets subordinating conjunctions like:
\begin{itemize}
	\item \textit{"that"} in "I know \textbf{that} he left"
	\item \textit{"whether"} in "Decide \textbf{whether} to go"
\end{itemize}

\textbf{Lexical Complexity Analysis}.
The lexical module evaluates vocabulary richness through two orthogonal measures:

\begin{equation}
	\text{Diversity} = \frac{|\mathcal{V}|}{N}, \quad 
	\text{Rarity} = \sum_{w \in \mathcal{W}} \mathbb{I}(|w| > \tau)
\end{equation}

We adopt the length-based rarity threshold $\tau=6$ to count the number of occurrences of low-frequency words.

\textbf{Feature Fusion Mechanism}.
The final complexity score combines syntactic and lexical features through manually set weights:

\begin{equation}
	C = \underbrace{0.3\alpha}_{\text{Clauses}} + \underbrace{0.2\beta}_{\text{Dependencies}} + \underbrace{0.3\gamma}_{\text{Rarity}} + \underbrace{0.2\delta}_{\text{Diversity}}
\end{equation}

\textbf{Complex Reasoning Decision}.
The final decision layer applies thresholding on the computed score shown in Equation \ref{reasoning decision}, where $\theta=0.65$.

\begin{equation}
	\text{Require Reasoning?} = 
	\begin{cases}
		\text{Yes} & \text{if } C > \theta \\
		\text{No} & \text{otherwise}
	\end{cases}
	\label{reasoning decision}
\end{equation}
\section{Prompt Engineering Details}\label{prompt details}

In this section, we provide the verbatim prompt templates utilized across our experimental settings to ensure reproducibility. We categorize these prompts into four distinct components:

\begin{itemize}[leftmargin=*]
	\item \textbf{Standard Inference (Baseline):} The zero-shot prompt used for the base model evaluation, where the model directly answers the question based on the video frames.
	\item \textbf{Explicit Visual Anchoring:} The specific prompt designed to force the model to generate a question-aware summary. This serves as the foundational "Step 1" in our proposed pipeline.
	\item \textbf{Naive Prompting Verification:} A comparative baseline where the model is simply asked to self-evaluate its previous answer without intermediate reasoning steps.
	\item \textbf{Latent Reasoning Consensus\:} The complete multi-turn dialogue template for our method. It integrates the \textit{Visual Anchoring} summary (Round 1) to drive \textit{Evidence-Grounded Deduction} (Round 2), leading to the \textit{Refined Response} (Round 3).
\end{itemize}

The specific templates are presented below.

\begin{tcolorbox}[
	colframe=black,
	colback=gray!10,
	coltitle=white,
	colbacktitle=black,
	fonttitle=\bfseries,
	title={Standard Inference Prompt (Baseline)},
	sharp corners=southwest,
	enhanced,
	breakable,
	center title,
	listing only,
	listing options={
		basicstyle=\ttfamily\small,
		breaklines=true,
	}
	]
	<|im\_start|>system
	
	You are a helpful assistant.<|im\_end|>
	
	<|im\_start|>user
	
	<image>
	
	The input consists of a sequence of key frames from a video. Please answer the following question:
	<Question>
	
	<im\_start>assistant
	
	<model\_output>
\end{tcolorbox}

\begin{tcolorbox}[
	colframe=black,
	colback=gray!10,
	coltitle=white,
	colbacktitle=black,
	fonttitle=\bfseries,
	title={Prompt for Explicit Visual Anchoring},
	sharp corners=southwest,
	enhanced,
	breakable,
	center title,
	listing only,
	listing options={
		basicstyle=\ttfamily\small,
		breaklines=true,
	}
	]
	<|im\_start|>system
	
	You are a helpful assistant.<|im\_end|>
	
	<|im\_start|>user
	
	<image>
	
	The input consists of a sequence of key frames from a video.
	
	Summarize the main content in the video, paying special attention to content related to the question:
	<Question>
	
	Content unrelated to the question can be summarized more briefly.
	<|im\_end|>
	
	<|im\_start|>assistant
	
	<summary\_output>
\end{tcolorbox}

\raggedbottom

\begin{tcolorbox}[
	colframe=black,
	colback=gray!10,
	fonttitle=\bfseries,
	title={Naive Prompting Verification -- Round 1: Initial Response},
	sharp corners=southwest,
	enhanced,
	before skip=10pt,
	after skip=10pt,
	center title,
	watermark color=gray!20,
	watermark text={Dialogue Structure}
	]
	\ttfamily
	<|im\_start|>system
	
	You are a helpful assistant.<|im\_end|>
	
	<|im\_start|>user
	
	<image>
	
	The input consists of a sequence of key frames from a video. Please answer the following question:
	<Question>
	
	<im\_start>assistant
	
	<round1\_output>
\end{tcolorbox}

\begin{tcolorbox}[
	colframe=black,
	colback=gray!10,
	fonttitle=\bfseries,
	title={Naive Prompting Verification -- Round 2: Naive Self Verify},
	sharp corners=southwest,
	enhanced,
	before skip=10pt,
	after skip=10pt,
	center title,
	watermark color=gray!20,
	watermark text={Dialogue Structure}
	]
	\ttfamily
	<|im\_start|>system
	
	You are a helpful assistant.<|im\_end|>
	
	<|im\_start|>user
	
	<image>
	
	The input consists of a sequence of key frames from a video.
	
	Please answer the following questions:
	<Question>
	
	Here is an answer to this question:
	<round1\_output>
	
	How reliable do you think this answer is?
	
	A. very reliable
	
	B. generally reliable 
	
	C. not very reliable 
	
	D. absolutely impossible
	
	Only select the best answer.
	
	<im\_start>assistant
	
	<round2\_output>
\end{tcolorbox}


\begin{tcolorbox}[
	colframe=black,
	colback=gray!10,
	fonttitle=\bfseries,
	title={Latent Reasoning Consensus -- Round 1: Explicit Visual Anchoring},
	sharp corners=southwest,
	enhanced,
	before skip=10pt,
	after skip=10pt,
	center title,
	watermark color=gray!20,
	watermark text={Dialogue Structure}
	]
	\ttfamily
	<|im\_start|>system
	
	You are a helpful assistant.<|im\_end|>
	
	<|im\_start|>user
	
	<image>
	
	The input consists of a sequence of key frames from a video.
	
	Summarize the main content in the video, paying special attention to content related to the question:<Question>
	
	Content unrelated to the question can be summarized more briefly.
	<|im\_end|>
	
	<|im\_start|>assistant
	
	<round1\_output>
\end{tcolorbox}


\begin{tcolorbox}[
	colframe=black,
	colback=gray!10,
	fonttitle=\bfseries,
	title={Latent Reasoning Consensus -- Round 2: Evidence-Grounded Deduction},
	sharp corners=southwest,
	enhanced,
	before skip=10pt,
	after skip=10pt,
	center title,
	watermark color=gray!20,
	watermark text={Dialogue Structure}
	]
	\ttfamily
	<|im\_start|>system
	
	You are a helpful assistant.<|im\_end|>
	
	<|im\_start|>user
	
	<image>
	
	The input consists of a sequence of key frames from a video.
	
	The main content of the video is summarized below:
	<round1\_output> 
	
	Here is a question about the video:
	<Question>
	
	Let's think step by step:<|im\_end|>
	
	<|im\_start|>assistant
	
	<round2\_output>
\end{tcolorbox}

\begin{tcolorbox}[
	colframe=black,
	colback=gray!10,
	fonttitle=\bfseries,
	title={Latent Reasoning Consensus -- Round 3: Refined Response},
	sharp corners=southwest,
	enhanced,
	before skip=10pt,
	after skip=10pt,
	center title,
	watermark color=gray!20,
	watermark text={Dialogue Structure}
	]
	\ttfamily
	<|im\_start|>system
	
	You are a helpful assistant.<|im\_end|>
	
	<|im\_start|>user
	
	<image>
	
	The input consists of a sequence of key frames from a video.
	
	The main content of the video is summarized below:
	<round1\_output> 
	
	Here is a question about the video:
	<Question>
	
	Let's think step by step:
	
	<round2\_output>
	
	Only select the best answer. The final answer is:
	<|im\_end|>
	
	<|im\_start|>assistant
	
	<round3\_output>
\end{tcolorbox}

\end{document}